\documentclass[sigconf]{acmart}
\usepackage{latexsym}
\usepackage{amsmath}
\usepackage{amsfonts}
\usepackage{booktabs}
\usepackage{multirow}
\usepackage{enumitem}
\usepackage{colortbl}
\usepackage{pifont}

\copyrightyear{2026}
\acmYear{2026}
\setcopyright{cc}
\setcctype{by}
\acmConference[CIKM '26] {Proceedings of the 35th ACM International Conference on Information and Knowledge Management}{November 7--11, 2026}{Rome, Italy.}
\acmBooktitle{Proceedings of the 35th ACM International Conference on Information and Knowledge Management (CIKM '26), November 7--11, 2026, Rome, Italy}
\acmISBN{979-8-4007-2539-5/2026/11}
\acmDOI{10.1145/3799682.3841148}

\begin{document}

\title{Neutralizing Popularity Bias in LLM-based Recommendation via Counterfactual Reasoning Guidelines}


\author{Guanrong Li}
\affiliation{%
  \institution{Nanjing University}
  \city{Nanjing}
  \state{Jiangsu}
  \country{China}
}
\email{grli@smail.nju.edu.cn}

\author{Haolin Yang}
\affiliation{%
  \institution{Nanjing University}
  \city{Nanjing}
  \state{Jiangsu}
  \country{China}
}
\email{haolinyang@smail.nju.edu.cn}

\author{Xinyu Liu}
\affiliation{%
  \institution{Nanjing University}
  \city{Nanjing}
  \state{Jiangsu}
  \country{China}
}
\email{xinyuliu@smail.nju.edu.cn}

\author{Zhen Wu}
\affiliation{%
  \institution{Nanjing University}
  \city{Nanjing}
  \state{Jiangsu}
  \country{China}
}
\email{wuz@nju.edu.cn}

\author{Rui Xia}
\affiliation{%
  \institution{Nanjing University}
  \city{Nanjing}
  \state{Jiangsu}
  \country{China}
}
\email{rxia@nju.edu.cn}

\author{Xinyu Dai}
\affiliation{%
  \institution{Nanjing University}
  \city{Nanjing}
  \state{Jiangsu}
  \country{China}
}
\email{daixinyu@nju.edu.cn}
\renewcommand{\shortauthors}{Guanrong Li et al.}

\begin{abstract}
In the era of generative AI, recommender systems are moving from precise prediction to trustworthy generation. Large language models (LLMs) support this shift by inferring user interests and producing natural-language explanations. However, LLM-based recommendation suffers from a fundamental obstacle: popularity bias. Through pre-training on massive corpora, LLMs tend to rely on global statistics and trend signals, yielding recommendations that follow popularity rather than genuine preference. As this bias is entangled in model parameters and is hard to remove directly, we propose \textbf{N}eutralizing \textbf{P}opularity Bias in LLM-based \textbf{Rec}ommendation via Counterfactual Reasoning Guidelines (NPRec), a model-agnostic framework that mitigates popularity bias through external semantic intervention. NPRec performs counterfactual refinement to causally separate intrinsic user interests from popularity-driven conformity, producing debiased textual guidelines that reflect actual user preferences. These guidelines are injected at inference time to shift the LLM from unconstrained generation to guided reasoning, without any parameter updates. Serving as explicit premises, they both ground faithful explanations and improve recommendation quality. Extensive experiments on three real-world datasets demonstrate that NPRec achieves promising performance in recommendation accuracy, explanation quality and debiasing capability.
\end{abstract}




\maketitle
\section{Introduction}
Recommender systems are undergoing a paradigm shift from traditional collaborative filtering toward generative, semantic-aware reasoning enabled by Large Language Models (LLMs) \cite{yang2024fine, xie2023factual, chu2024llm}. Unlike classical models that rely on rigid item-user overlaps, LLMs can project historical behaviors into a unified semantic space, enabling not only precise interest inference but also natural-language explanations that bridge recommendation and user trust \cite{bao2023tallrec, geng2022recommendation}.

A fundamental yet overlooked obstacle is the semantic popularity bias inherent in LLMs. While traditional popularity bias stems from skewed interaction data, LLM-based recommenders can further amplify such bias through popularity priors encoded during pre-training, over-preferring frequent entities (e.g., global blockbusters) embedded in their massive training corpora, often overriding the user's specific, long-tail interests \cite{wang2025does, lichtenberg2024large}. This creates a faithfulness gap: a fluent explanation is merely a ``hallucinated justification'' if the recommendation is driven by internal frequency priors rather than the provided user evidence.

Existing solutions do not fully address this problem. Classical popularity-debiasing methods are mainly designed for ID-based recommenders and do not transfer naturally to free-form LLM generation \cite{SchnabelSSCJ16, liu2024interact}. More recent LLM-based methods often rely on fine-tuning with bias-aware data, which is computationally costly and embeds the debiasing effect into updated model parameters, making the intervention implicit and harder to interpret, verify, and connect to faithful explanations \cite{lu2025dual, luo2026bridging}. Prompting and in-context learning are also limited, because they still rely on the same biased model to correct itself without explicitly separating user-specific evidence from popularity priors.

To address this problem, we propose NPRec (Neutralizing Popularity Bias in LLM-based Recommendation), a model-agnostic framework that mitigates popularity bias through external semantic intervention. NPRec first synthesizes user interaction patterns into explicit textual guidelines. It then evaluates each guideline under two causal scenarios: a factual setting with full user-item signals and an auxiliary counterfactual setting in which user-specific variables are replaced by uninformative reference states. By subtracting the counterfactual score from the factual estimate, NPRec prioritizes guidelines supported by user-specific evidence rather than popularity-driven conformity. The refined guidelines are finally injected into the LLM as explicit premises for recommendation, shifting debiasing from implicit parameter adjustment to explicit counterfactual-guided reasoning.

This framework offers three strategic advantages for robust recommendation. First, it enables inference-time intervention without modifying the underlying LLM parameters. This model-agnostic design avoids additional training of the underlying LLM, ensures computational efficiency, and provides operational transparency that is unattainable in "black-box" fine-tuning approaches. Second, NPRec replaces unconstrained generation with evidence-grounded reasoning. By forcing the model to operate on counterfactually refined guidelines, it effectively bypasses the semantic shortcuts triggered by popularity signals during recommendation. Third, NPRec significantly improves explanation faithfulness. In NPRec, the refined guidelines are directly used as the explanations themselves, while also serving as the semantic premises for recommendation. As a result, the explanations are true reflections of the model's inference process, rather than being post-hoc rationalizations.

Experiments on three real-world datasets show that NPRec improves recommendation quality, debiasing performance and explanation quality. We further study a setting with human-written guidelines, which suggests that better guideline quality can further improve the system. In addition, NPRec supports a hybrid deployment paradigm in which high-capacity LLMs are used offline for guideline generation, while smaller LLMs are used online for guided inference, making the framework more efficient for practical use. The main contributions are summarized as follows:
\begin{itemize}[leftmargin=*, nosep]
\item We propose NPRec, a model-agnostic framework for LLM-based recommendation that mitigates popularity bias through \textbf{external semantic intervention}. By moving debiasing from parameter adjustment to explicit guideline-based control, NPRec introduces a new paradigm for popularity-robust LLM recommendation.
\item We introduce a \textbf{counterfactual-guided reasoning} mechanism that refines candidate guidelines by subtracting popularity-driven counterfactual effects from factual preference estimates. This original design performs debiasing directly in the semantic reasoning space, rather than in the parameter space, while also improving transparency of the decision.
\item Extensive experiments on three real-world datasets demonstrate that NPRec improves debiasing capability, recommendation accuracy, and explanation quality. Further ablation and guideline studies confirm the effectiveness of the proposed design and show that it naturally supports hybrid deployment, where strong LLMs are used for offline guideline generation and smaller LLMs are used for efficient online inference.
\end{itemize}

\section{Related Work}
\subsection{Large Language Models for Recommendation}
Large Language Models (LLMs) have been increasingly used for recommendation because they can jointly model textualized user histories, item information, and preference patterns \cite{lin2025can, zhao2024recommender}. Existing methods mainly follow two directions. One adapts LLMs through supervised or instruction tuning on recommendation data \cite{tsai2024leveraging, kim2408review, fang2025large, ferraretto2023exaranker, kang2023llms}, while the other keeps LLMs frozen and uses prompting, reasoning, or agent-based pipelines to infer user interests \cite{wang2024can, zhang2024agentcf}. The former may absorb popularity patterns into model parameters, whereas the latter still depends on popularity-related semantic priors learned during pre-training.
Recent studies have begun to examine popularity bias in LLM-based recommendation \cite{lichtenberg2024large, lu2025dual, luo2026bridging}. Existing methods are often diagnostic, rely on additional training, or ask the biased LLM itself to revise its predictions. In contrast, NPRec performs debiasing outside the LLM by generating and counterfactually refining textual guidelines before final recommendation.

\subsection{Popularity Debiasing in Recommendation}
Popularity bias refers to the tendency of recommender systems to over-recommend popular items while under-exposing less frequent but still relevant ones. It has long been studied in conventional ID-based recommendation. Representative solutions include inverse propensity scoring and exposure reweighting \cite{SchnabelSSCJ16, tan2021counterfactual}, causal or disentangled modeling of preference and exposure \cite{zheng2021disentangling}, and post-hoc re-ranking for fairness or diversity control \cite{abdollahpouri2019managing}. These methods are mainly designed for score-based pipelines, where user-item relevance is explicitly modeled and debiasing is applied to ranking scores or exposure distributions.
LLM-based recommendation creates a different setting. Popularity bias can arise not only from interaction statistics, but also from the semantic priors of pre-trained LLMs, which may affect the generated reasoning process. Therefore, conventional debiasing methods cannot be directly transferred. Some recent LLM-oriented methods attempt to reduce this issue \cite{lu2025dual, gao2025sprec}, but they often require extra fine-tuning or auxiliary training objectives, making the debiasing effect implicit in model parameters.
NPRec instead debiases the semantic evidence provided to the LLM. It first externalizes candidate preference descriptions as textual guidelines, and then refines them by subtracting a popularity-driven counterfactual component from the factual guideline score. In this way, debiasing is performed in the semantic reasoning space rather than the parameter space. The refined guidelines serve as debiased control signals, helping the LLM rely more on user-specific evidence and less on popularity shortcuts.

\subsection{Natural Language Explanations for Recommendation}
Natural-language explanations can improve the transparency and usability of recommender systems. With the development of LLMs, recent work has increasingly studied free-form explanations for recommendation \cite{yang2024fine, xie2023factual, chu2024llm, li2023personalized, wang2022coffee}. These methods can generate fluent and informative text, but the explanation is often post-hoc: it is produced after the recommendation has been made and may not reflect the evidence that actually influenced the decision.
This leads to a faithfulness gap, where the explanation may be plausible but mainly acts as a rationalization. NPRec addresses this issue by using refined guidelines as part of the recommendation process itself. This design creates a tighter connection between what guides the decision and what is shown to the user.

\section{Task Formulation and Causal Analysis}\label{sec:theoretical}
We first formalize the recommendation task and the causal intuition behind NPRec. The recommendation results can be influenced by both user-specific evidence and popularity-driven conformity. NPRec is designed to preserve the former while reducing the latter.

\subsection{Task Formalization}
Let $\mathcal{U} = \{u_1, u_2, \dots, u_m\}$ and $\mathcal{I} = \{i_1, i_2, \dots, i_n\}$ denote the sets of users and items, respectively. For each user $u \in \mathcal{U}$, let $\mathcal{H}_u = \{i_1, i_2, \dots, i_{|\mathcal{H}_u|}\}$ denote the interaction history and let $\mathcal{C}_u \subset \mathcal{I}$ denote the candidate item set. Our goal is to generate both a recommendation $\hat{i}$ and a corresponding natural-language explanation $e_u$:
\begin{equation}
(\hat{i}, e_u) = f(u, \mathcal{H}_u, \mathcal{C}_u),
\end{equation}
where $f(\cdot)$ denotes the overall recommendation framework. The key challenge is that, in LLM-based recommendation, the generated reasoning may be influenced not only by the user's actual interests, but also by popularity cues learned from large-scale corpora. Therefore, beyond prediction accuracy, we seek a recommendation process in which the decision is driven as much as possible by user-specific preference rather than popularity-driven conformity. We formalize this requirement in the following causal analysis.

\begin{figure}[!t]
  \centering
  \includegraphics[width=0.75\linewidth]{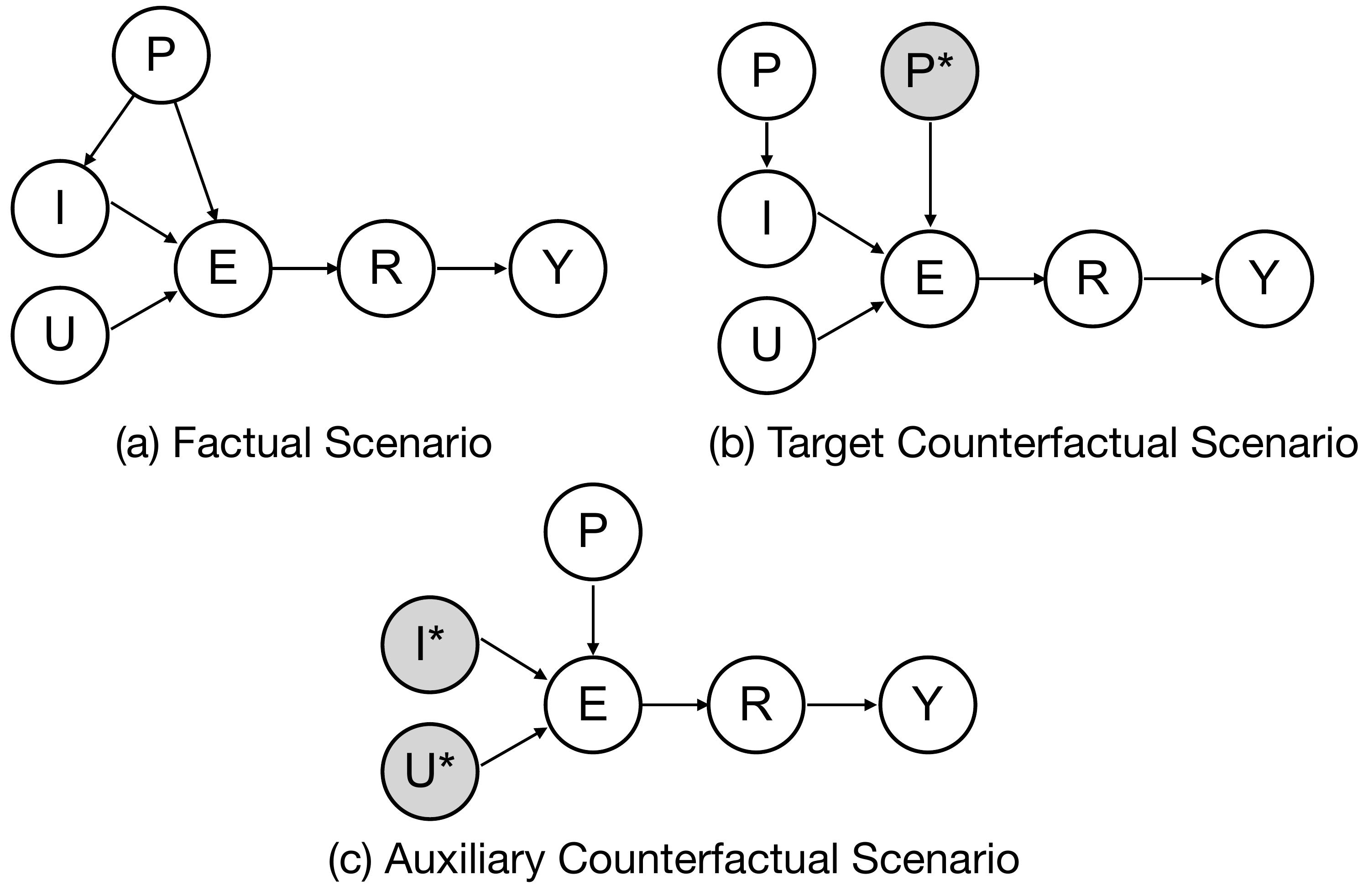}
  \vspace{-0.75em}
\caption{Causal visualization of the counterfactual debiasing. (a) Factual Scenario: The observed SCM where popularity ($P$) influences guidelines ($E$) through two pathways: an indirect path ($P \to I \to E$) that reflects exposure-related item signals, and a direct path ($P \to E$) that reflects conformity bias. (b) Target Counterfactual Scenario: The ideal inference state where the direct conformity path is removed while exposure-related signals are preserved. (c) Auxiliary Counterfactual Scenario: A proxy configuration used to estimate the popularity-driven component by intervening on $U$ and $I$ to suppress user-specific paths.}
  \label{causal_graph}
\end{figure}

\begin{figure*}[!t]
  \centering
  \includegraphics[width=0.8\textwidth]{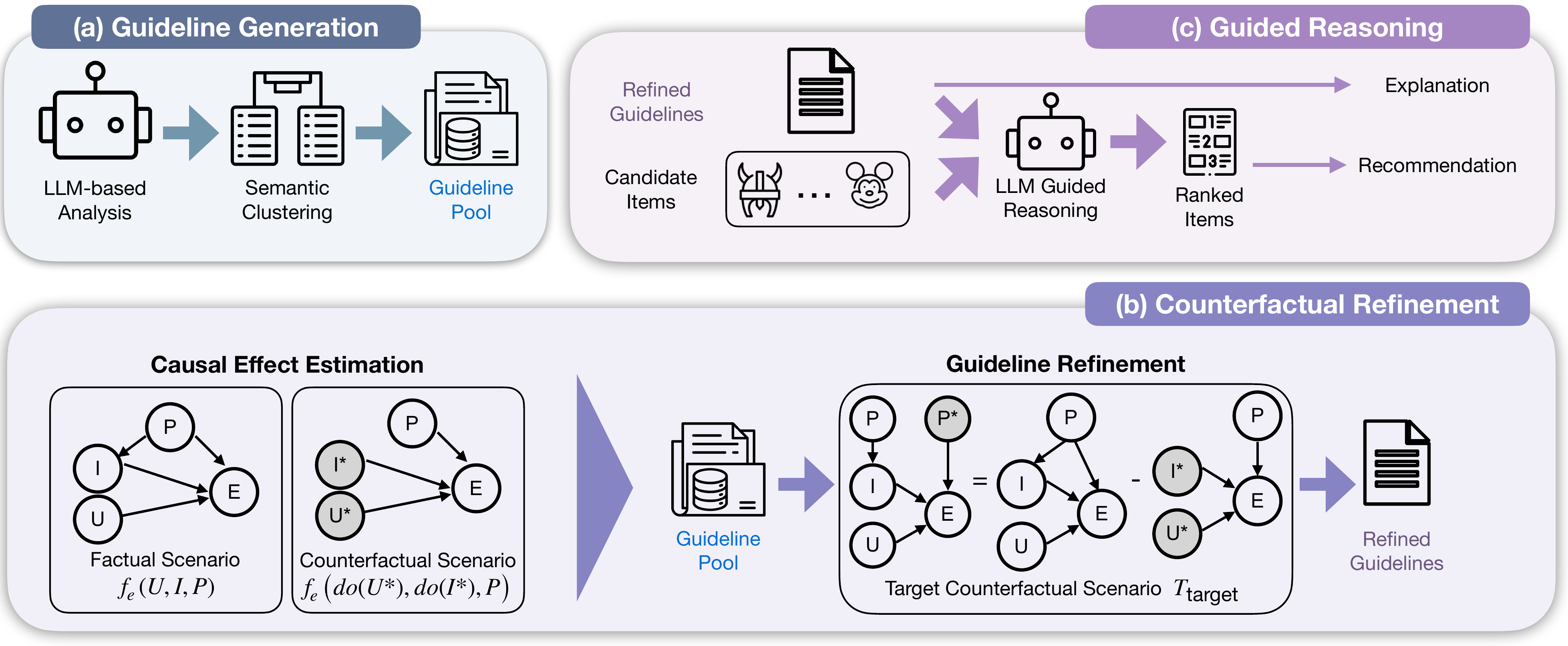}
    \vspace{-0.75em}
  \caption{Overview of the NPRec framework. The framework consists of three stages: (a) Guideline Generation, which converts user-item interaction data into candidate semantic guidelines; (b) Counterfactual Refinement, which estimates the factual score $f_e(u, i, p)$ and the auxiliary counterfactual score $f_e(do(u^*), do(i^*), p)$, and then uses their difference to refine the guidelines; and (c) Guided Reasoning, where an LLM performs recommendation based on the refined guidelines and returns them as explanations.}
  \label{model_structure}
\end{figure*}

\subsection{Causal View of Popularity Bias}\label{sec:causal_view}
To describe how popularity affects LLM-based recommendation, we model the guideline generation process with a Structural Causal Model (SCM), shown in Figure \ref{causal_graph}(a). We define the endogenous variables as
\(\mathbf{V} = \{U, I, P, E, R, Y\},\)
where $U$ denotes the user profile, $I$ denotes item-side signals available to the model, $P$ denotes item popularity, $E$ denotes the generated guideline, $R$ denotes the LLM reasoning process, and $Y$ denotes the final recommendation result.

The central variable in our analysis is the guidelines $E$. It is influenced by user information, item-related signals, and popularity. Ideally, the path $\{U, I\} \to E$ captures user-specific preference where the model generates a guideline because it is supported by the user’s history and by the candidate item. However, popularity may also affect guidelines $E$ through two different routes.

\begin{itemize}[leftmargin=*]
\item \textbf{Exposure-related indirect path $P \to I \to E$.} Popularity can shape what kinds of items are more likely to be encountered, interacted with, or represented in the user’s observed history. In this sense, some popularity influence is part of the user’s factual exposure and should not be removed completely. In our SCM, this is represented by the indirect path through item-side signals $I$.

\item \textbf{Conformity path $P \to E$.} Popularity can also directly bias the guideline, even when the user's own evidence does not support it. This path captures the tendency of an LLM to over-activate concepts simply because they are globally frequent or widely discussed. This is the popularity effect that we want to reduce.
\end{itemize}

Our target is therefore not to erase all traces of popularity. Instead, we aim to reduce the part of guideline relevance that can be explained by popularity under reference user-item states, while retaining guidelines supported by factual user-item evidence.

\subsection{Counterfactual Debiasing Strategy}
Based on the SCM above, our goal is to favor guidelines supported by user-specific evidence while reducing guideline activation that can be explained by popularity under non-informative user-item conditions. Let $E$ denote the guideline variable and let $e$ be a candidate guideline. Under the factual user-item-popularity context, guideline relevance can be written as
$q_{\theta}(e \mid u, i, p)$
where $u$, $i$, and $p$ denote the user, item-side signal, and item popularity, respectively.
Since popularity-driven guideline activation cannot be directly observed, we introduce an auxiliary reference intervention. Specifically, we replace the user and item inputs with reference states drawn from $\rho_U$ and $\rho_I$, while keeping the popularity signal unchanged:
\begin{equation}
q_{\theta}^{\mathrm{ref}}(e \mid p)
=
\mathbb{E}_{\tilde{u}\sim \rho_U, \tilde{i}\sim \rho_I}
\left[
q_{\theta}(e \mid do(U=\tilde{u}), do(I=\tilde{i}), p)
\right].
\end{equation}
This reference distribution estimates how likely guideline $e$ is to be activated when user-specific and item-specific evidence is suppressed. We then define the desired intervention as
\begin{equation} 
\label{m-e}
T_{\mathrm{target}}(u,i,p,e) =
q_{\theta}(e \mid u, i, p) - q_{\theta}^{\mathrm{ref}}(e \mid p).
\end{equation}
A larger $T_{\mathrm{target}}$ indicates that guideline $e$ is better supported by the factual user-item context and less explainable by popularity.

\section{Methodology}\label{sec:methodology}
\subsection{Framework Overview}
As illustrated in Figure \ref{model_structure}, NPRec follows a three-stage workflow to mitigate popularity bias: \textbf{Guideline Generation} (Figure \ref{model_structure}(a)): In this step, we use an offline LLM to analyze raw interaction logs, generate explicit semantic descriptions (\textit{guidelines}), and construct a candidate guideline pool. As LLM generation is guided by learned token probabilities from large-scale corpora, the resulting guidelines inherit popularity bias. Therefore, this stage externalizes the bias into text, setting the stage for the causal intervention. \textbf{Counterfactual Refinement} (Figure \ref{model_structure}(b)): This step operationalizes our debiasing logic. We employ counterfactual inference to disentangle genuine user preference signals from popularity-driven conformity. Specifically, we estimate the likelihood of a guideline being triggered in a counterfactual scenario where the user and item representations are replaced by universal reference states. By subtracting this popularity-driven component from the factual estimate, we filter out guidelines that merely reflect statistical popularity, ensuring that the retained guidelines represent the user's intrinsic interests in a causal sense. \textbf{Guided Reasoning} (Figure \ref{model_structure}(c)): In this step, the LLM performs reasoning strictly conditioned on the unbiased, refined guidelines. By grounding inference in these counterfactually debiased signals, we reduce LLMs' reliance on popularity-based shortcuts. This not only ensures \textit{explanatory faithfulness}, because the guidelines are the actual antecedents of the decision, but also improves recommendation accuracy by forcing the model to align its reasoning with the user’s intrinsic interests.

\subsection{Guideline Generation}
As shown in Figure \ref{model_structure}(a), the Guideline Generation stage transforms user interaction data into semantic patterns by leveraging the LLM’s synthesis capabilities. Instead of treating histories as ID sequences, we prompt the LLM to identify recurring themes from a user's history $\mathcal{H}_u$ and their target item $i^*$. To enhance robustness, we apply sub-sequence augmentation, truncating histories at random timestamps to create multiple interest snapshots. This process is formalized as:
$\mathcal{G}_u = \text{LLM}_\phi (u, \mathcal{H}_u, i^*)$,
where $\mathcal{G}_u$ is the guideline set and $\text{LLM}_\phi$ is the inference procedure.

To capture global patterns, we aggregate guidelines into a pool $\mathcal{G} = \text{Cluster} \left( \bigcup_{u \in \mathcal{U}} \mathcal{G}_u \right)$ through semantic clustering. This denoises individual user quirks and condenses the guideline space into representative behavioral prototypes. We use MiniLM embeddings \cite{wang2021minilmv2} and HDBSCAN \cite{campello2015hierarchical} to cluster the guidelines, enabling semantic alignment across users even without shared items. It is important to note that the guideline pool is constructed offline using only the training split. During evaluation, NPRec does not generate guidelines from validation or test interactions.

\subsection{Counterfactual Refinement}
As illustrated in Figure \ref{model_structure}(b), this stage operationalizes our debiasing logic via causal effect estimation, extracting guidelines that are both behaviorally relevant and statistically unbiased. To compute the target causal effect in Eq. \ref{m-e}, we estimate two components: (i) the probability that a guideline is triggered under the joint influence of the user’s genuine preference and popularity (the factual scenario in Figure \ref{causal_graph}(a)), and (ii) the probability that it is triggered by popularity conformity alone after intervening to remove user-specific signals (the counterfactual scenario in Figure \ref{causal_graph}(c)). To this end, we first construct a robust popularity representation that serves as the key input for causal modeling. We then introduce the causal estimator architectures for both the factual and counterfactual scenarios, together with the optimization procedure used to train them. Finally, we describe the guideline refinement mechanism, which leverages these learned causal effect estimates to filter out popularity bias.

\textbf{(1) Popularity Representation.}
We approximate item popularity using interaction frequency in the training data. Since item frequencies are highly skewed, we apply logarithmic scaling and min-max normalization to obtain a normalized popularity score $n_i$ for each item $i$. Directly using $n_i$ as a scalar may be too limited to model the non-linear effect of popularity on user preference. We therefore project it into a dense popularity representation $\mathbf{p}_i$ through soft bucket interpolation.
Specifically, we divide the normalized popularity range into $K$ overlapping buckets. Let $start_k$ and $end_k$ denote the boundaries of the $k$-th bucket, and let $\mathbf{v}^b_k$ be its learnable bucket embedding. For item $i$, we compute its interpolation weight for bucket $k$ as $w_k(n_i)=\frac{\max(0,\min(n_i-start_k,end_k-n_i))}{end_k-start_k}$, and obtain the final popularity representation as $\mathbf{p}_i=\sum_{k=1}^{K}w_k(n_i)\mathbf{v}^b_k$. Compared with hard bucket assignment, this soft interpolation avoids abrupt changes at bucket boundaries and gives similar representations to items with similar popularity scores, while still allowing popularity to be represented in a higher-dimensional space.

\textbf{(2) Causal Effect in Factual Scenario.}
In this section, we define the estimation function $f_e(u,i,p)$ for the factual scenario. Drawing inspiration from neural network-based collaborative filtering, we integrate Generalized Matrix Factorization (GMF) and a Deep Feature Cross (DFC) architecture \cite{he2017neural, cheng2016wide}, specifically modified to incorporate item popularity. For a specific guideline $g$, the estimated effect for a given user $u$ and item $i$ is:
\begin{equation}\label{eq:gmf_p} 
f_P^{G}(u,i,p) = \text{MLP}^{G}\left([\mathbf{u} \odot \mathbf{i} ; \mathbf{p}_i]\right) 
\end{equation}
\begin{equation}\label{eq:dfc_p}
    f_p^{D}(u,i,p) = \text{MLP}^{D}\left([\text{MLP}(\mathbf{u}, \mathbf{i}) ; \mathbf{p}_i]\right) 
\end{equation}
\begin{equation}\label{eq:factual_f}
\hat{f}^g_e(u,i,p)  = \sigma\left(\left[ f_P^{G}(u,i,p) \odot \mathbf{g}^{G} ; \text{MLP}\left( [f_p^{D}(u,i,p) ; \mathbf{g}^{D}] \right) \right] \right)
\end{equation}
Here $\sigma(\cdot)$ denotes the sigmoid function, and $[\cdot ; \cdot]$ represents the concatenation operation. $\mathbf{u}, \mathbf{i}$, $\mathbf{p}_i$ and $\mathbf{g}$ are the embedding vectors for the user, item, item popularity and guideline, respectively. $f_P^{G}$ captures element-wise product interactions typical of matrix factorization, while $f_p^{D}$ utilizes a multi-layer perceptron to model high-order, non-linear feature crosses. The separation of guideline representation into $\mathbf{g}^{G}$ and $\mathbf{g}^{D}$ allows for tailoring the guideline's influence to the distinct structural properties of the GMF and DFC modules.

\textbf{(3) Causal Effect in Auxiliary Counterfactual Scenario.} Here we estimate the causal effect in the auxiliary counterfactual scenario, $f_e(do(U^*), do(I^*), P)$, where both the user and item are intervened upon and set to a universal reference state.
\begin{equation}\label{eq:counter_aux}
\hat{f}^g_e(do(u^*),do(i^*),p) = \sigma\left(\left[ \mathbf{r}^G \odot \mathbf{g}^{G} ; \text{MLP} \left( [ \mathbf{r}^D ; \mathbf{g}^{D} ] \right) \right] \right)
\end{equation}
In practice, we simplify the reference distributions $\rho_U$ and $\rho_I$ by using deterministic zero-vector reference states $u^*$ and $i^*$.
This null reference state removes user-specific and item-specific signals from the estimator input, so that the post-intervention representations $\mathbf{r}^G$ and $\mathbf{r}^D$ are constructed solely from the popularity representation $\mathbf{p}_i$ through learnable transformations. Accordingly, this operationally suppresses user-specific and item-specific contributions along the $U \rightarrow E$ and $I \rightarrow E$ paths in the auxiliary scenario. Therefore, $\hat{f}^g_e(do(u^*),do(i^*),p)$ serves as an approximation to the popularity-driven component, with minimal interference from user-specific preference signals.

\textbf{(4) Estimation Optimization.} Following the formulation of causal effects in both the factual and auxiliary counterfactual scenarios, we optimize the estimation parameters through a multi-task learning objective. Rather than computing exact probabilities across the entire guideline space, which is computationally prohibitive given the massive scale of the candidate pool, we optimize these parameters through a sampled contrastive multi-task learning objective. Specifically, for a given interaction $(u,i)$, the model aims to distinguish the ground-truth guideline $g$ from a set of negative candidates $\mathcal{G}_g^-$ in both scenarios:
\begin{equation} 
\mathcal{L}_{\text{factual}} = - \sum_{g \in \mathcal{G}_{u,i}} \log \frac{\exp(\hat{f}^g_e(u,i,p))}{ \sum\limits_{g' \in \{g\} \cup \mathcal{G}_g^-} \exp(\hat{f}^{g'}_e(u,i,p))}
\end{equation}
\begin{equation}
    \mathcal{L}_{\text{aux}} = - \sum_{g \in \mathcal{G}_{u,i}} \log \frac{\exp(\hat{f}_e^g(do(u^*),do(i^*),p))}{ \sum\limits_{g' \in \{g\} \cup \mathcal{G}_g^-} \exp(\hat{f}_e^{g'}(do(u^*),do(i^*),p))}
\end{equation}
\begin{equation} \label{eq:optimization}
\mathcal{L} = \mathcal{L}_{\text{factual}} + \alpha \mathcal{L}_{\text{aux}} 
\end{equation}
The first term, $\mathcal{L}_{\text{factual}}$, supervises the model to learn the comprehensive causal relationships present in the factual scenario. The second term, $\mathcal{L}_{\text{aux}}$, specifically trains the model to capture the popularity-driven causal effect. $\hat{f}_e(u,i,p)$ and $\hat{f}_e(do(u^*),do(i^*),p)$ are shown in Equations \ref{eq:factual_f} and \ref{eq:counter_aux}, respectively. The hyperparameter $\alpha$ serves as a weight to balance these two objectives. 

\textbf{(5) Guideline Refinement.}
As illustrated in the right half of Figure \ref{model_structure}(b), we perform inference based on the theoretical framework in Equation \ref{m-e}. We calculate a debiased matching score by subtracting the popularity-driven causal effect (auxiliary counterfactual scenario) from the total estimated effect (factual scenario):
\begin{equation}\label{beta_coff}
t_{\text{debiased}}(u,i,g) = \hat{f}^g_e(u,i,p) - \beta \hat{f}^g_e(do(u^*),do(i^*),p)
\end{equation}
Here, the factual term $\hat{f}^g_e$ represents the composite signal of user interest and popularity conformity, while the counterfactual term isolates the pure popularity bias. The hyperparameter $\beta$ controls the intensity of this bias subtraction. This operation effectively filters out the conformity effect, leaving a residual score that reflects the user's legitimate, intrinsic interest. To identify the debiased guidelines that best represent preference for a user $u$ across candidate items $\mathcal{C}_u$, we retrieve the top-\(N\) guidelines \(\hat{\mathcal G}_u\) from the global pool \(\mathcal G\):
\begin{equation}\label{eq:TopN}
\hat{\mathcal G}_u= \text{Top-N}_{g \in \mathcal{G}, i_c \in \mathcal{C}_u} t_{\text{debiased}}(u, i_c, g)
\end{equation}
By maximizing the debiased score rather than the raw probability, this retrieval step ensures that the retrieved guidelines better reflect genuine user preferences rather than generic popularity trends.

\subsection{Guided Reasoning}
As shown in Figure \ref{model_structure}(c), following the retrieval of debiased guidelines, we utilize the LLM to reason about the recommendation task. This process adheres to the causal chain $E \to R \to Y$, where the guidelines ($E$) inform the reasoning process ($R$), which in turn determines the final recommendation outcome ($Y$). Within this framework, the guidelines serve as natural, interpretable explanations for the recommendation results. The reasoning process is formalized as:
\begin{equation}
\hat{i} = f_{\text{reason}}(u, \mathcal{H}_u, \mathcal{C}_u, \hat{\mathcal G}_u)
\end{equation}
where $f_{\text{reason}}$ denotes the LLM reasoning procedure. Given the user $u$, their interaction history $\mathcal{H}_u$, a candidate item set $\mathcal{C}_u$, and the specialized guidelines $\hat{\mathcal G}_u$, the model outputs the recommended item \(\hat{i}\) and directly uses the guidelines \(\hat{\mathcal G}_u\) as the explanation. By conditioning the final inference on counterfactually refined guidelines, the LLM is shielded from popularity-driven shortcuts, ensuring the generated recommendations are grounded in personalized logic.

\section{Experiments}
In this section, we first set up the experiments and conduct extensive experiments to address the following research questions:
\begin{itemize}[leftmargin=*, nosep]
    \item \textbf{RQ1:} Can NPRec jointly improve recommendation performance, explanation quality, and debiasing capability compared with existing recommendation and LLM-based baselines?
    \item \textbf{RQ2:}  Which components are critical to NPRec, and how do LLM backbones, hybrid deployment settings, and guideline quality, including human-written guidelines, influence the final recommendation performance?
    \item \textbf{RQ3:} What kinds of guidelines does NPRec retrieve, and do these refined guidelines genuinely steer the LLM's final recommendation behavior rather than merely serve as post-hoc explanations?
\end{itemize}

\begin{table}[tbp]\small
  \centering
  \caption{Statistical summary of datasets. Ave. indicates average.}
          \vspace{-0.75em}
    \begin{tabular}{lccc}
    \toprule
      Datasets    & MovieLens & CDs and Vinyl & Books\\
    \midrule
    Number of users & 604   & 467 & 935\\
    Number of items  & 1863  & 1627 & 2906\\
    Number of interactions & 64760 & 8591 & 25278\\
    Avg. interactions per user & 107.40 & 18.44 & 27.06\\
    Avg. interactions per item & 34.78 & 5.28 & 8.70\\
    Sparsity & 94.24 & 98.87 & 99.07\\
    \bottomrule
    \end{tabular}
  \label{tab:datasets}%
\end{table}%

\subsection{Experimental Setup}
\textbf{Datasets.}
We evaluate NPRec on three real-world datasets: MovieLens, a movie recommendation benchmark with user--movie interactions and metadata, and two subsets of the Amazon Reviews dataset \cite{ni2019justifying}, namely \textit{CDs and Vinyl} and \textit{Books}. To ensure fair comparison across different LLM backbones and baselines, we apply a unified preprocessing pipeline. Specifically, we randomly sample 1,000 users from Amazon Books and 500 users from Amazon CDs and Vinyl. Across all datasets, we retain only users and items with at least three interactions. For each user, we keep at most 10 historical interactions. During evaluation, we construct a candidate set of 20 items, including 2 positive items and 18 randomly sampled negative items. Detailed dataset statistics are provided in Table \ref{tab:datasets}.

\begin{table}[t]
  \centering
  \caption{Capabilities of different baselines.}
    \vspace{-0.75em}
  \resizebox{\linewidth}{!}{
    \begin{tabular}{l|ccccc}
    \toprule
    \multicolumn{1}{c|}{Models} & Recommendation & Explanation & Explanation Format & Training-free LLMs & Debias \\
    \midrule
    NCF   & \ding{52} & \ding{55} & None  & -     & \ding{55} \\
    SimpleX & \ding{52} & \ding{55} & None  & -     & \ding{55} \\
    LightGCN & \ding{52} & \ding{55} & None  & -     & \ding{55} \\
    \midrule
    IPS   & \ding{52} & \ding{55} & None  & -     & \ding{52} \\
    DICE  & \ding{52} & \ding{55} & None  & -     & \ding{52} \\
    CERS  & \ding{52} & \ding{55} & None  & -     & \ding{52} \\
    CountER & \ding{52} & \ding{52} & Keywords & -     & \ding{52} \\
    FGCR  & \ding{52} & \ding{52} & Keywords & -     & \ding{52} \\
    \midrule
    PEPLER & \ding{55} & \ding{52} & Natural Language & \ding{55} & \ding{55} \\
    XRec  & \ding{55} & \ding{52} & Natural Language & \ding{55} & \ding{55} \\
    \midrule
    Prompting & \ding{52} & \ding{52} & Natural Language & \ding{52} & \ding{55} \\
    CoT   & \ding{52} & \ding{52} & Natural Language & \ding{52} & \ding{55} \\
    D\textsuperscript{2}LR  & \ding{52} & \ding{55} & None  & \ding{55} & \ding{52} \\
    SPRec & \ding{52} & \ding{55} & None  & \ding{55} & \ding{52} \\
    NPRec (ours) & \ding{52} & \ding{52} & Natural Language & \ding{52} & \ding{52} \\
    \bottomrule
    \end{tabular}}
  \label{tab:baselinescomperation}%
\end{table}

\textbf{Baselines.}
We compare NPRec with four groups of baselines. The first group includes standard recommendation models: NCF~\cite{he2017neural}, SimpleX~\cite{mao2021simplex}, and LightGCN~\cite{he2020lightgcn}. These focus on learning user-item matching signals but do not address explanation generation.

The second group includes popularity-debiasing and counterfactual recommendation methods: IPS~\cite{SchnabelSSCJ16}, CERS~\cite{liu2024interact}, DICE~\cite{zheng2021disentangling}, CountER~\cite{tan2021counterfactual}, and FGCR~\cite{xia2023toward}. These baselines reduce popularity bias or provide counterfactual explanations through reweighting, causal modeling, or explanation-aware recommendation. 

The third group includes LLM-based debiasing methods, namely $D^2LR$~\cite{lu2025dual} and SPRec~\cite{gao2025sprec}, which mitigate popularity bias through LLM training or self-play. The fourth group includes explainable recommendation methods, PEPLER~\cite{li2023personalized} and XRec~\cite{ma2024xrec}, which generate natural-language explanations for recommendation results.

We also compare with direct LLM prompting and chain-of-thought prompting~\cite{wei2022chain} to evaluate whether NPRec improves over standard LLM reasoning baselines.

\begin{table*}[h]\small
  \centering
\caption{Performance comparison of different methods. The best results for each LLM backbone are highlighted in \textbf{bold}.}
  \vspace{-0.75em}
\begin{tabular}{l|c|ccc|c|ccc|c|ccc}
\toprule
\multicolumn{1}{c|}{\textbf{Models}} & \multicolumn{4}{c|}{\textbf{MovieLens}} & \multicolumn{4}{c|}{\textbf{CDs and Vinyl}} & \multicolumn{4}{c}{\textbf{Books}} \\
      & \textbf{Top-1} & \multicolumn{3}{c|}{\textbf{Top-2}} & \textbf{Top-1} & \multicolumn{3}{c|}{\textbf{Top-2}} & \textbf{Top-1} & \multicolumn{3}{c}{\textbf{Top-2}} \\
      & \textbf{Hit} & \textbf{Hit} & \textbf{Recall} & \textbf{NDCG} & \textbf{Hit} & \textbf{Hit} & \textbf{Recall} & \textbf{NDCG} & \textbf{Hit} & \textbf{Hit} & \textbf{Recall} & \textbf{NDCG} \\
\midrule
NCF   & 0.3282 & 0.4788 & 0.3568 & 0.3420 & 0.3014 & 0.4444 & 0.3535 & 0.3107 & 0.2532 & 0.4699 & 0.3245 & 0.3010 \\
SimpleX & 0.3399 & 0.5390 & 0.3516 & 0.3469 & 0.3197 & 0.4210 & 0.3755 & 0.3202 & 0.2550 & 0.4595 & 0.3419 & 0.3033 \\
LightGCN & 0.3217 & 0.5506 & 0.3524 & 0.3410 & 0.2803 & 0.4859 & 0.3242 & 0.3117 & 0.2620 & 0.4442 & 0.3202 & 0.3049 \\
\midrule
IPS   & 0.0978 & 0.1493 & 0.1012 & 0.0822 & 0.2039 & 0.2973 & 0.2666 & 0.1949 & 0.1680 & 0.2444 & 0.2027 & 0.1527 \\
DICE  & 0.3400 & 0.5290 & 0.3507 & 0.3555 & 0.2915 & 0.3898 & 0.2953 & 0.2883 & 0.2551 & 0.4301 & 0.3178 & 0.3094 \\
CERS  & 0.3367 & 0.4992 & 0.3601 & 0.3423 & 0.2907 & 0.4475 & 0.3743 & 0.2904 & 0.2497 & 0.4771 & 0.3263 & 0.3104 \\
CountER & 0.1905 & 0.3464 & 0.1843 & 0.1989 & 0.1708 & 0.2350 & 0.2121 & 0.2118 & 0.1987 & 0.2998 & 0.1898 & 0.1747 \\
FGCR  & 0.1776 & 0.3570 & 0.2140 & 0.2017 & 0.1822 & 0.2439 & 0.2389 & 0.2179 & 0.2003 & 0.3353 & 0.2107 & 0.2045 \\
\midrule
Qwen3.5 4B+Prompting & 0.1355 & 0.2556 & 0.2233 & 0.1832 & 0.1804 & 0.3059 & 0.2631 & 0.1880 & 0.2069 & 0.3128 & 0.2660 & 0.2407 \\
Qwen3.5 4B+CoT & 0.1552 & 0.2833 & 0.2365 & 0.2054 & 0.1833 & 0.3083 & 0.2625 & 0.1833 & 0.2010 & 0.3226 & 0.2692 & 0.2417 \\
Qwen3.5 4B+D\textsuperscript{2}LR & 0.1906 & 0.2748 & 0.2215 & 0.2064 & 0.2162 & 0.3636 & 0.3059 & 0.2712 & 0.2669 & 0.3728 & 0.2660 & 0.2975 \\
Qwen3.5 4B+SPRec & 0.1892 & 0.2998 & 0.2568 & 0.2299 & 0.2039 & 0.3243 & 0.2801 & 0.2509 & 0.2877 & 0.4133 & 0.3172 & 0.3001 \\
Qwen3.5 4B+NPRec & \textbf{0.2020} & \textbf{0.3177} & \textbf{0.2623} & \textbf{0.2374} & \textbf{0.2246} & \textbf{0.3811} & \textbf{0.3185} & \textbf{0.2803} & \textbf{0.3031} & \textbf{0.4391} & \textbf{0.3240} & \textbf{0.3087} \\
\midrule
LLaMA-2-7B+Prompting & 0.1117 & 0.2256 & 0.1585 & 0.1273 & 0.1206 & 0.2206 & 0.1292 & 0.1073 & 0.1879 & 0.2453 & 0.1876 & 0.1759 \\
LLaMA-2-7B+CoT & 0.1208 & 0.2237 & 0.1698 & 0.1432 & 0.1189 & 0.1429 & 0.1479 & 0.1127 & 0.1877 & 0.2573 & 0.1838 & 0.1617 \\
LLaMA-2-7B+D\textsuperscript{2}LR & 0.1543 & 0.2865 & 0.2287 & 0.2012 & 0.1579 & 0.2514 & 0.2332 & 0.2142 & 0.2167 & 0.3350 & 0.2406 & 0.1999 \\
LLaMA-2-7B+SPRec & 0.1508 & 0.2778 & \textbf{0.2407} & 0.2068 & 0.1562 & 0.2581 & 0.2121 & 0.2043 & 0.2236 & \textbf{0.3366} & 0.2534 & 0.2006 \\
LLaMA-2-7B+NPRec & \textbf{0.1769} & \textbf{0.2976} & 0.2141 & \textbf{0.2202} & \textbf{0.1621} & \textbf{0.2693} & \textbf{0.2359} & \textbf{0.2153} & \textbf{0.2406} & 0.3333 & \textbf{0.2543} & \textbf{0.2186} \\
\midrule
GPT-3.5+Prompting & 0.2371 & 0.3284 & 0.1915 & 0.2004 & 0.2039 & 0.3243 & 0.2801 & 0.2509 & 0.2674 & 0.4123 & 0.2939 & 0.2849 \\
GPT-3.5+CoT & 0.2571 & 0.4013 & 0.2430 & 0.2434 & 0.2285 & 0.3710 & 0.3145 & 0.2746 & 0.2808 & 0.4243 & 0.3053 & 0.2988 \\
GPT-3.5+NPRec & \textbf{0.3632} & \textbf{0.5406} & \textbf{0.3466} & \textbf{0.3470} & \textbf{0.2604} & \textbf{0.3759} & \textbf{0.3231} & \textbf{0.2990} & \textbf{0.3085} & \textbf{0.4480} & \textbf{0.3292} & \textbf{0.3135} \\
\bottomrule
\end{tabular}%
  \label{tab:recommend}%
\end{table*}%

\begin{table*}[htbp]\small
  \centering
  \caption{Explanation performance across different models. Just measures how well the explanation justifies the recommendation results, while Con measures consistency with historical user reviews. Evaluations are conducted using LLM-based evaluation ($\text{Just}_\text{LLM}$, $\text{Con}_\text{LLM}$) and human evaluation ($\text{Just}_\text{Human}$, $\text{Con}_\text{Human}$). Best results are in \textbf{bold}.}
    \vspace{-0.75em}
\resizebox{\textwidth}{!}{
\begin{tabular}{rrrrrrrrrrrr}
\toprule
\multicolumn{1}{c|}{\multirow{2}[2]{*}{Models}} & \multicolumn{1}{c|}{\multirow{2}[2]{*}{Formats}} & \multicolumn{2}{c|}{MovieLens} & \multicolumn{4}{c|}{CDs and Vinyl } & \multicolumn{4}{c}{Books} \\
\multicolumn{1}{c|}{} & \multicolumn{1}{c|}{} & \multicolumn{1}{c}{Just\textsubscript{LLM} } & \multicolumn{1}{c|}{Just\textsubscript{Human}} & \multicolumn{1}{c}{Just\textsubscript{LLM}} & \multicolumn{1}{c}{Just\textsubscript{Human}} & \multicolumn{1}{c}{Con\textsubscript{LLM}} & \multicolumn{1}{c|}{Con\textsubscript{Human}} & \multicolumn{1}{c}{Just\textsubscript{LLM}} & \multicolumn{1}{c}{Just\textsubscript{Human}} & \multicolumn{1}{c}{Con\textsubscript{LLM}} & \multicolumn{1}{c}{Con\textsubscript{Human}} \\
\midrule
\multicolumn{1}{l|}{CountER} & \multicolumn{1}{c|}{Keywords} & \multicolumn{1}{c}{0.1111} & \multicolumn{1}{c|}{0.12} & \multicolumn{1}{c}{0.6592} & \multicolumn{1}{c}{0.42} & \multicolumn{1}{c}{0.5202} & \multicolumn{1}{c|}{0.24} & \multicolumn{1}{c}{0.5147} & \multicolumn{1}{c}{0.14} & \multicolumn{1}{c}{0.5364} & \multicolumn{1}{c}{0.18} \\
\multicolumn{1}{l|}{FGCR} & \multicolumn{1}{c|}{Keywords} & \multicolumn{1}{c}{0.1447} & \multicolumn{1}{c|}{0.16} & \multicolumn{1}{c}{0.6472} & \multicolumn{1}{c}{0.40} & \multicolumn{1}{c}{0.5466} & \multicolumn{1}{c|}{0.32} & \multicolumn{1}{c}{0.5569} & \multicolumn{1}{c}{0.10} & \multicolumn{1}{c}{0.5971} & \multicolumn{1}{c}{0.20} \\
\midrule
\multicolumn{1}{l|}{PEPLER} & \multicolumn{1}{c|}{Natural Language} & \multicolumn{1}{c}{0.7482} & \multicolumn{1}{c|}{0.50} & \multicolumn{1}{c}{0.6773} & \multicolumn{1}{c}{0.58} & \multicolumn{1}{c}{0.5621} & \multicolumn{1}{c|}{0.44} & \multicolumn{1}{c}{0.6896} & \multicolumn{1}{c}{0.62} & \multicolumn{1}{c}{0.5656} & \multicolumn{1}{c}{0.50} \\
\multicolumn{1}{l|}{XRec} & \multicolumn{1}{c|}{Natural Language} & \multicolumn{1}{c}{0.7477} & \multicolumn{1}{c|}{0.58} & \multicolumn{1}{c}{\textbf{0.6924}} & \multicolumn{1}{c}{0.72} & \multicolumn{1}{c}{0.5983} & \multicolumn{1}{c|}{0.50} & \multicolumn{1}{c}{0.6847} & \multicolumn{1}{c}{0.72} & \multicolumn{1}{c}{0.5725} & \multicolumn{1}{c}{\textbf{0.52}} \\
\multicolumn{1}{l|}{GPT-3.5+Prompting} & \multicolumn{1}{c|}{Natural Language} & \multicolumn{1}{c}{0.7215} & \multicolumn{1}{c|}{0.32} & \multicolumn{1}{c}{0.6201} & \multicolumn{1}{c}{0.58} & \multicolumn{1}{c}{0.5193} & \multicolumn{1}{c|}{0.48} & \multicolumn{1}{c}{0.5181} & \multicolumn{1}{c}{0.54} & \multicolumn{1}{c}{0.5337} & \multicolumn{1}{c}{0.32} \\
\multicolumn{1}{l|}{GPT-3.5+CoT} & \multicolumn{1}{c|}{Natural Language} & \multicolumn{1}{c}{0.7390} & \multicolumn{1}{c|}{0.38} & \multicolumn{1}{c}{0.6241} & \multicolumn{1}{c}{0.62} & \multicolumn{1}{c}{0.5117} & \multicolumn{1}{c|}{0.44} & \multicolumn{1}{c}{0.6764} & \multicolumn{1}{c}{0.70} & \multicolumn{1}{c}{0.5192} & \multicolumn{1}{c}{0.44} \\
\multicolumn{1}{l|}{GPT-3.5+NPRec} & \multicolumn{1}{c|}{Natural Language} & \multicolumn{1}{c}{\textbf{0.7662}} & \multicolumn{1}{c|}{\textbf{0.62}} &  \multicolumn{1}{c}{0.6854} & \multicolumn{1}{c}{\textbf{0.74}} & \multicolumn{1}{c}{\textbf{0.6000}} & \multicolumn{1}{c|}{\textbf{0.52}} & \multicolumn{1}{c}{\textbf{0.7171}} & \multicolumn{1}{c}{\textbf{0.78}} & \multicolumn{1}{c}{\textbf{0.6459}} & \multicolumn{1}{c}{\textbf{0.52}} \\
\bottomrule
\end{tabular}}
  \label{tab:explain}%
\end{table*}

\noindent\textbf{Capability Comparison.}
Table \ref{tab:baselinescomperation} summarizes the functional differences among the compared methods. Traditional recommenders may improve ranking quality or reduce popularity bias, but they generally do not provide fluent natural-language explanations. Existing LLM-based methods can generate readable text, but many either require additional training or remain vulnerable to popularity bias. In contrast, NPRec is designed to jointly support recommendation, natural-language explanation, training-free LLM inference, and popularity debiasing within a unified framework.

\noindent\textbf{Evaluation Metrics.}
For recommendation quality, we report Hit Ratio (Hit), Recall, and Normalized Discounted Cumulative Gain (NDCG). For explanation quality, we evaluate two dimensions: \textit{Justification}, which measures whether the explanation supports the recommended item, and \textit{Consistency}, which measures whether the explanation agrees with the user's historical reviews. We conduct human evaluation on 50 randomly sampled cases and apply an LLM-based evaluator to the full test set. Each explanation is scored on a three-point scale, where $-1$, $0$, and $1$ indicate conflict, partial agreement, and full agreement, respectively.

To measure whether a method reduces popularity-oriented exposure, we report two exposure-based debiasing metrics: Average Recommendation Popularity (ARP@K) and Long-tail Coverage (LTCov@K). Let $I$ be the item set, and let $c_i$ be the interaction frequency of item $i$ in the training set. We define the popularity score of item $i$ as $\phi(i)=\log(1+c_i)$, where a larger $\phi(i)$ indicates higher popularity. For each user $u \in U$, let $\mathrm{Rec}_K(u)$ denote the top-$K$ recommended items. ARP@K is defined as $\mathrm{ARP@K}=\frac{1}{|U|K}\sum_{u\in U}\sum_{i\in \mathrm{Rec}_K(u)}\phi(i)$, which measures the average popularity of exposed items. A lower ARP@K indicates less exposure to highly popular items.
For long-tail exposure, let $I_{\text{tail}}\subseteq I$ denote the bottom 50\% of items ranked by interaction frequency. LTCov@K is defined as $\mathrm{LTCov@K}=\frac{\left|\left(\bigcup_{u\in U}\mathrm{Rec}_K(u)\right)\cap I_{\text{tail}}\right|}{|I_{\text{tail}}|}$, which measures the proportion of distinct long-tail items exposed in the top-$K$ recommendations across all users. A higher LTCov@K indicates wider exposure of long-tail items.

\noindent\textbf{Implementation Details.}
Our implementation is based on well-established packages, including Hugging Face Transformers \cite{wolf-etal-2020-transformers} and RecBole \cite{recbole[1.2.0]}. All experiments are conducted on 2 NVIDIA Tesla V100 GPUs, and the code is available on GitHub.\footnote{https://github.com/kylokano/NPRec} We repeat each experiment three times with different random seeds and report the average results.
For automated evaluation of explanation quality, we use Gemini 2.5 Pro as the LLM-based evaluator. For human evaluation, each result is independently annotated by two annotators, and we report the average score across annotators. Hyperparameters are tuned separately for each dataset. Specifically, the configuration of $\{\alpha,\beta,N\}$ is set to $\{0.01,1,1\}$ for Amazon CDs, $\{0.01,1,10\}$ for Amazon Books, and $\{0.01,5,5\}$ for MovieLens, where $\alpha$ controls the balance of the loss function in Equation~\ref{eq:optimization}, $\beta$ controls the strength of the counterfactual intervention in Equation~\ref{beta_coff}, and $N$ denotes the number of retrieved guidelines in Equation~\ref{eq:TopN}. Due to the high sparsity of the Amazon CDs and Vinyl dataset, approximately 40\% of the results produced by CountER and FGCR do not have explanations. Therefore, for these two baselines, we compute explanation-related metrics only on the subset of results with available explanations. To keep training efficient, we train D\textsuperscript{2}LR and SPRec with LoRA \cite{Hu2021LoRALA} using rank $r=8$.

\begin{table*}[htbp]\small
  \centering
  \caption{Debiasing performance on popular and long-tail items. We report Recall and NDCG on three datasets, with \textit{Popularity} and \textit{Long-Tail} referring to the top 15\% and bottom 50\% of items by interaction frequency, respectively.}
    \vspace{-0.75em}
\begin{tabular}{l|cc|cc|cc|cc|cc|cc}
\toprule
\multirow{3}[2]{*}{Models} & \multicolumn{4}{c|}{MovieLens} & \multicolumn{4}{c|}{CDs and Vinyl} & \multicolumn{4}{c}{Books} \\
      & \multicolumn{2}{c|}{Popularity} & \multicolumn{2}{c|}{Long-Tail} & \multicolumn{2}{c|}{Popularity} & \multicolumn{2}{c|}{Long-Tail} & \multicolumn{2}{c|}{Popularity} & \multicolumn{2}{c}{Long-Tail} \\
      & Recall & NDCG  & Recall & NDCG  & Recall & NDCG  & Recall & NDCG  & Recall & NDCG  & Recall & NDCG \\
\midrule
IPS   & 0.1012 & 0.0876 & 0.0539 & 0.0355 & 0.2614 & 0.1838 & 0.2654 & 0.1847 & 0.1796 & 0.1329 & 0.1926 & 0.1403 \\
DICE  & 0.3588 & 0.3616 & 0.2835 & 0.1805 & 0.3555 & 0.3431 & 0.2515 & 0.2084 & 0.4220 & 0.3711 & 0.2942 & 0.2414 \\
CERS  & 0.3521 & 0.3578 & 0.2627 & 0.1629 & 0.4274 & 0.3653 & 0.2695 & 0.2308 & 0.4604 & 0.3971 & 0.2989 & 0.2541 \\
CountER & 0.2169 & 0.2246 & 0.1666 & 0.1675 & 0.2431 & 0.2341 & 0.2066 & 0.2069 & 0.2631 & 0.2046 & 0.1641 & 0.1572 \\
FGCR  & 0.2673 & 0.2552 & 0.1928 & 0.1525 & 0.2566 & 0.2467 & 0.2186 & 0.2089 & 0.2757 & 0.2520 & 0.1988 & 0.1833 \\
\midrule
\multicolumn{1}{l|}{Qwen3.5 4B+D\textsuperscript{2}LR} & 0.2491 & 0.2276 & 0.1576 & 0.1428 & 0.3201 & 0.2809 & 0.2244 & 0.2137 & 0.3102 & 0.3105 & 0.2546 & 0.2237 \\
\multicolumn{1}{l|}{Qwen3.5 4B+SPRec} & 0.2597 & 0.2491 & 0.2413 & 0.1718 & 0.2956 & 0.2627 & 0.2612 & 0.2315 & 0.3323 & 0.3152 & 0.2794 & 0.2431 \\
\multicolumn{1}{l|}{Qwen3.5 4B+NPRec} & 0.2644 & 0.2383 & 0.2459 & 0.1736 & 0.3239 & 0.2965 & 0.2585 & 0.2276 & 0.3444 & 0.3170 & 0.3023 & 0.2559 \\
GPT-3.5+NPRec & 0.3525 & 0.3704 & 0.2667 & 0.1767 & 0.4318 & 0.3755 & 0.2654 & 0.2294 & 0.4426 & 0.3991 & 0.2998 & 0.2654 \\
\bottomrule
\end{tabular}%
  \label{tab:fulldebias}
\end{table*}

\begin{table}[t]
  \centering
  \caption{Exposure-based debiasing results. \textit{ARP} measures the average popularity of the recommended item, where lower values indicate less reliance on popular items. \textit{LTCov} measures the proportion of distinct long-tail items appearing in recommendations, where higher values indicate broader exposure to long-tail items. LLM-based methods use Qwen3.5 4B as the backbone.}
  \vspace{-0.75em}
  \resizebox{\linewidth}{!}{
    \begin{tabular}{l|cc|cc|cc}
    \toprule
    \multirow{2}[2]{*}{Models} & \multicolumn{2}{c|}{MovieLens} & \multicolumn{2}{c|}{CDs and Vinyl} & \multicolumn{2}{c}{Books} \\
          & ARP@1 & LTCov@1 & ARP@1 & LTCov@1 & ARP@1 & LTCov@1 \\
    \midrule
    IPS   & 3.151 & 0.278 & 1.615 & 0.182 & 2.248 & 0.217 \\
    DICE  & 4.560 & 0.183 & 2.047 & 0.155 & 2.788 & 0.168 \\
    \midrule
    \multicolumn{1}{l|}{D\textsuperscript{2}LR} & 2.976 & 0.214 & 1.513 & 0.207 & 1.865 & 0.213 \\
    \multicolumn{1}{l|}{SPRec} & 3.393 & 0.279 & 1.641 & 0.202 & 1.983 & 0.197 \\
    \multicolumn{1}{l|}{NPRec} & 3.492 & 0.272 & 1.964 & 0.237 & 2.179 & 0.224 \\
    \bottomrule
    \end{tabular}}
  \label{tab:exposure_bias}
\end{table}

\subsection{Overall Performance (RQ1)}
Table \ref{tab:recommend} reports the recommendation results on three datasets. NPRec consistently outperforms all compared LLM-based baselines under the same backbone. In particular, it achieves clear gains over both direct prompting and CoT, showing that the improvement comes not simply from longer reasoning traces, but from the counterfactual refinement mechanism. The gains are especially clear on Top-1 recommendation, suggesting that the refined guidelines help the model focus on the most user-relevant item instead of generic or popularity-driven cues. NPRec is also competitive with strong traditional recommenders, and surpasses them in several settings, while additionally providing natural-language explanations.

For explanation quality, Table \ref{tab:explain} shows that NPRec achieves the best overall results across all datasets. Due to space limitations, for LLM-based prompting methods, we report only results using the GPT-3.5 backbone.
The results show that NPRec produces explanations that better justify the recommendation outcomes and align more closely with user-side preference evidence. This suggests that the counterfactually refined guidelines capture more genuine user interests by filtering out popularity-driven noise.

To evaluate debiasing ability, we further perform a stratified analysis over different popularity groups, where \textit{Popular} items denote the top 15\% by interaction frequency and \textit{Long-tail} items denote the bottom 50\%. As shown in Table \ref{tab:fulldebias}, NPRec maintains strong performance on popular items while also achieving competitive gains on long-tail items. In contrast, most baselines perform substantially worse on long-tail items, revealing stronger sensitivity to popularity bias. Existing debiasing methods partly improve long-tail performance, but often at the cost of overall recommendation quality. NPRec instead achieves a better balance, improving long-tail performance without sacrificing strong recommendation quality.

To further examine whether the final exposure is less concentrated on popular items, we report two exposure-based metrics, ARP@1 and Long-tail Coverage@1, in Table \ref{tab:exposure_bias}. The results show that NPRec reduces popularity-oriented exposure compared with standard LLM-based baselines while maintaining competitive recommendation quality. Although some specialized debiasing methods achieve stronger bias reduction on individual metrics, NPRec provides a more balanced trade-off among recommendation accuracy, explanation quality, and debiasing performance in the LLM-based recommendation setting.

\begin{table}[t]\small
  \centering
  \caption{Results of the ablation study. Here, w/o PD denotes removing Popularity Debiasing, w/ RG denotes using Random Guidelines, and w/o Cluster denotes removing Semantic Clustering.}
    \vspace{-0.75em}
  \resizebox{\linewidth}{!}{
\begin{tabular}{l|cc|cc|cc}
\toprule
      & \multicolumn{2}{c|}{MovieLens} & \multicolumn{2}{c|}{CDs and Vinyl} & \multicolumn{2}{c}{Books} \\
      & Recall & NDCG  & Recall & NDCG  & Recall & NDCG \\
\midrule
GPT-3.5+NPRec & 0.3466 & 0.3470 & 0.3231 & 0.2990 & 0.3292 & 0.3135 \\
w/o PD & 0.2529 & 0.2507 & 0.3145 & 0.2803 & 0.3090 & 0.3018 \\
w/ RG & 0.1930 & 0.1950 & 0.1551 & 0.1385 & 0.1429 & 0.1222 \\
w/o Cluster & 0.1863 & 0.1749 & 0.1613 & 0.1386 & 0.1414 & 0.1228 \\
\bottomrule
\end{tabular}}
  \label{tab:ablation}%
\end{table}

\subsection{Important Components (RQ2)}
Here we analyze the core components of NPRec from four perspectives: (1) core component ablations, (2) different LLM backbones and hybrid deployment settings, (3) hyperparameter sensitivity, and (4) the impact of human-written guidelines.

\textbf{(1) Ablation Study.}
We compare NPRec with three variants: \textit{w/o Popularity Debiasing}, which removes counterfactual refinement and directly uses the factual score $\hat{f}_e(u,i,p)$; \textit{w/ Random Guidelines}, which replaces retrieved user-specific guidelines with random ones from the global pool; and \textit{w/o Semantic Clustering}, which uses raw guidelines without clustering. As shown in Table \ref{tab:ablation}, all three variants perform worse than the full model. Removing popularity debiasing causes a clear drop, showing the importance of counterfactual refinement in filtering popularity-driven signals. Replacing retrieved guidelines with random ones leads to the largest degradation, indicating that the gain comes from meaningful semantic guidance rather than simply adding extra text to the prompt. Removing semantic clustering also hurts performance, suggesting that clustering helps transform noisy user-level descriptions into more stable and transferable prototypes.

\textbf{(2) Influence of LLM Backbones and Hybrid Deployment.}
Since NPRec separates \textit{guideline generation} from \textit{guided reasoning}, we further examine its behavior under different LLM backbones and hybrid settings. As shown in Table \ref{tab:LLM}, the framework is highly sensitive to guideline quality. When smaller models such as LLaMA-2-7B generate guidelines by themselves, performance is limited. In contrast, when stronger guidelines pre-generated by GPT-3.5 are provided, these smaller models achieve much better reasoning performance. This reveals a clear capability split: smaller LLMs can effectively follow high-quality semantic guidance, but are less reliable at synthesizing accurate guidelines from raw interaction history. This result supports the practical value of NPRec's hybrid design, where stronger LLMs generate guidelines offline and smaller models perform efficient online inference.

\begin{figure}[!t]
  \centering
  \includegraphics[width=\linewidth]{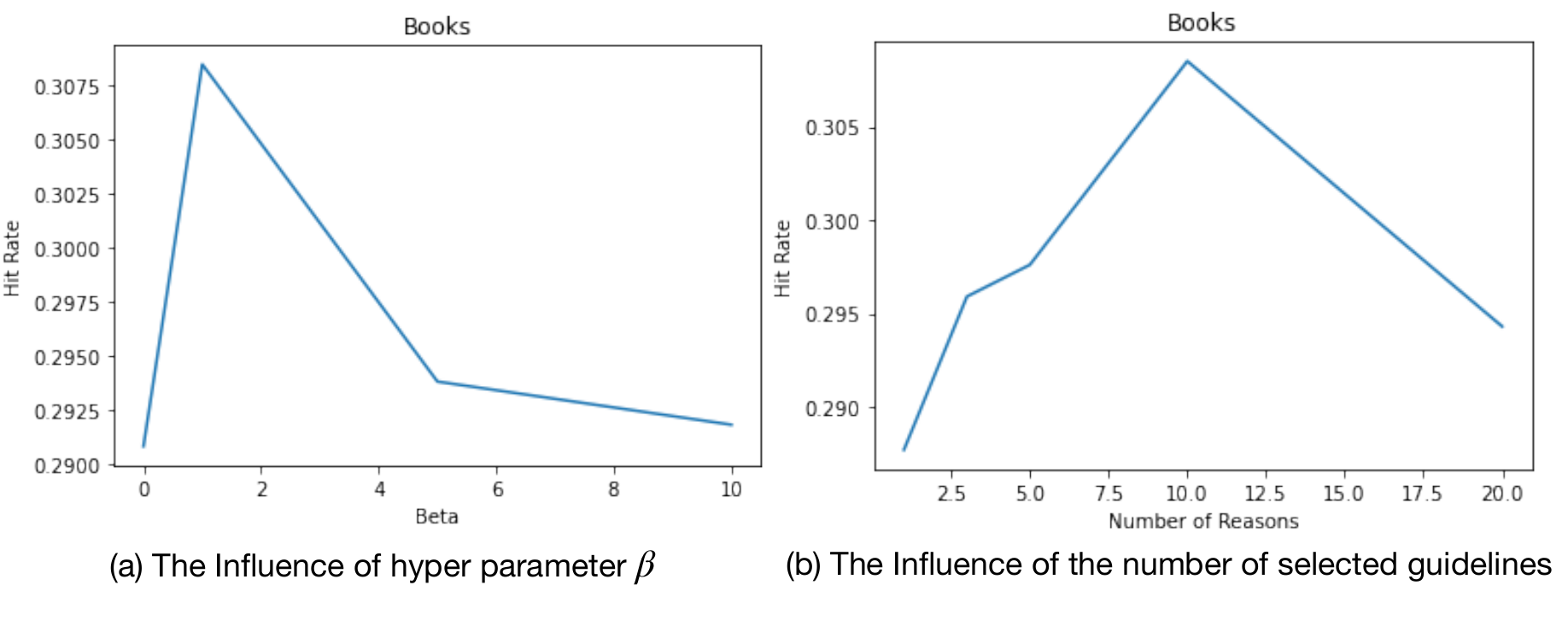}
        \vspace{-1em}
\caption{The influence of hyperparameters.}
  \label{pic:hyperparameters}
\end{figure}

\textbf{(3) Hyperparameter Sensitivity.} We investigate the sensitivity of NPRec to two key parameters: the debiasing coefficient $\beta$ (Eq. \ref{beta_coff}) and the number of guidelines retrieved via Counterfactual Refinement. For brevity, we present the results for the Amazon Books dataset in Figure \ref{pic:hyperparameters}. Our analysis identifies $\beta$ as a pivotal regulator of the framework's debiasing intensity. An optimal $\beta$ successfully mitigates the conformity effect without over-suppressing the underlying collaborative signals essential for accurate recommendation. Furthermore, we observe a sweet spot regarding the quantity of retrieved guidelines. An insufficient number of guidelines provides inadequate semantic context for the LLM's reasoning engine, whereas an excessive number introduces redundant information and semantic noise, potentially leading to reasoning distraction. Maintaining this delicate equilibrium is crucial for maximizing both the predictive accuracy and the explanatory fidelity.

\begin{figure}[!t]
  \centering
  \includegraphics[width=\linewidth]{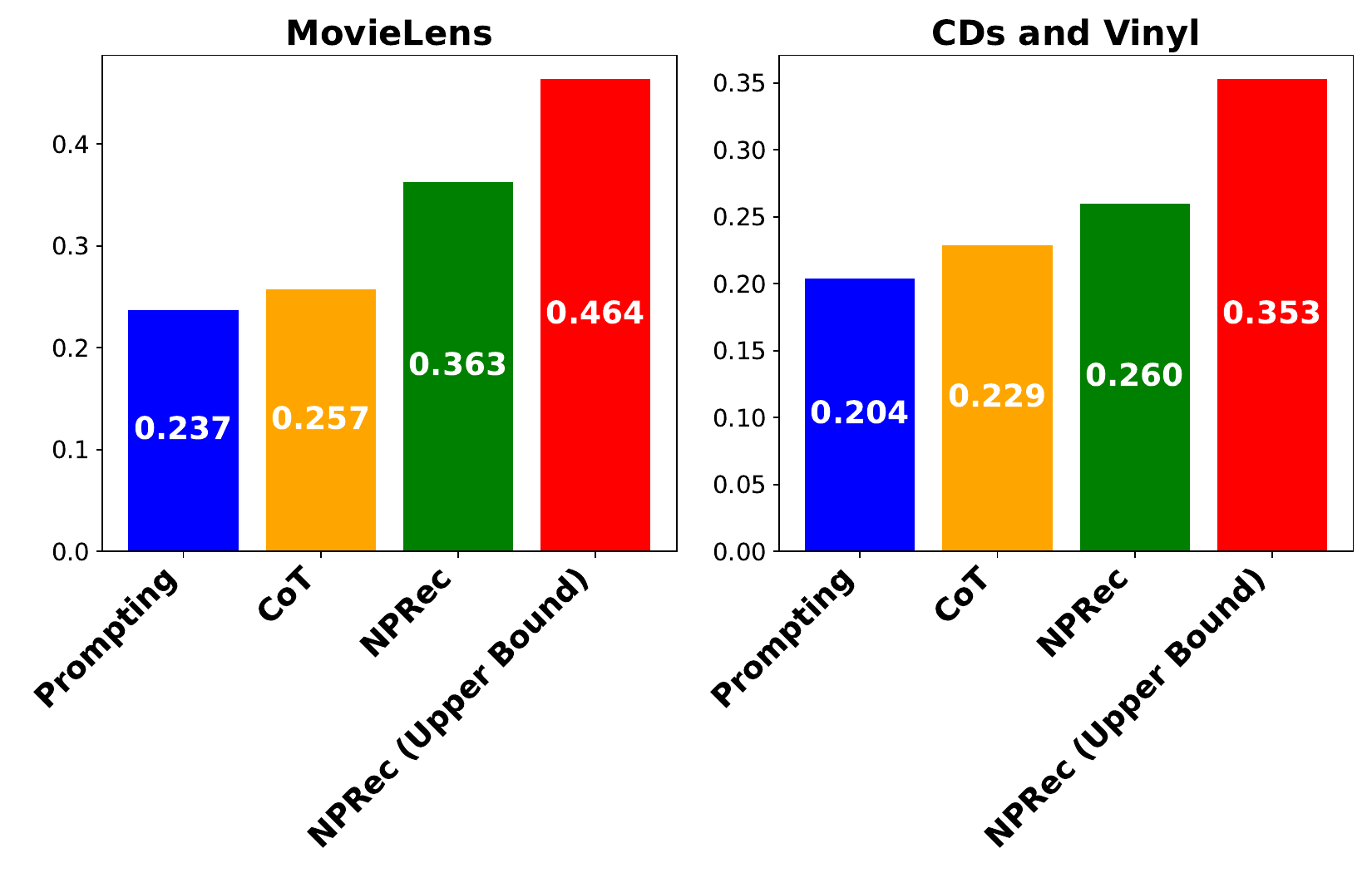}
\caption{Top-1 Hit results of NPRec across two datasets. These findings highlight the transformative potential of guided reasoning for LLM-based recommendation. We define the `Upper Bound' as the performance achieved using manually created guidelines on a subset of 100 samples. This comparison demonstrates a substantial elevation in the performance ceiling compared to basic prompting strategies, revealing gains of +95.6\% on MovieLens and +73.1\% on CDs and Vinyl.}
        \vspace{-0.75em}
  \label{upperbound}
\end{figure}

\textbf{(4) Human-written Guidelines.}
To further understand the role of guideline quality in NPRec, we conduct an additional study using human-written guidelines. The goal is to probe the performance headroom of the framework under higher-quality semantic guidance. 

For this analysis, we manually construct guidelines for a subset of 100 samples and use them in place of the automatically retrieved guidelines during the final reasoning stage. We then compare the resulting recommendation performance with standard prompting-based inference and with the default NPRec pipeline. Since this experiment requires manual annotation, we report it only as a small-scale diagnostic study. Figure \ref{upperbound} summarizes the Top-1 Hit results. We observe that manually written guidelines consistently lead to stronger recommendation performance across datasets. This result supports two conclusions. First, the quality of the guidelines is a key factor in the effectiveness of NPRec. The final recommendation quality depends not only on the reasoning ability of the LLM, but also on whether the provided guideline accurately captures the user-specific preference signal. Second, the gap between the default NPRec results and the manually written guideline results indicates that the current framework still has clear room for improvement in guideline generation and refinement.

\begin{table}[t]
  \centering
\caption{Performance across different LLM backbones. NPRec\textsubscript{Base} denotes the end-to-end framework using a unified model for both generation and reasoning. NPRec\textsubscript{Hybrid} represents a hybrid deployment setting where the model performs reasoning based on a guideline pool pre-synthesized by GPT-3.5.}
  \vspace{-0.75em}
\resizebox{\columnwidth}{!}{
\begin{tabular}{l|cc|cc|cc}
\toprule
\multicolumn{1}{c|}{\multirow{2}[2]{*}{Models}} & \multicolumn{2}{c|}{MovieLens} & \multicolumn{2}{c|}{CDs and Vinyl} & \multicolumn{2}{c}{Books} \\
      & Recall & NDCG  & Recall & NDCG  & Recall & NDCG \\
\midrule
LLaMA-2-7B & 0.1585 & 0.1273 & 0.1292 & 0.1073 & 0.1876 & 0.1759 \\
NPRec\textsubscript{Base} & 0.2141 & 0.2202 & 0.2359 & 0.2153 & 0.2543 & 0.2186 \\
NPRec\textsubscript{Hybrid} & 0.2554 & 0.2332 & 0.2741 & 0.2464 & 0.2982 & 0.2749 \\
\midrule
LLaMA-2-13B & 0.1573 & 0.1280 & 0.1309 & 0.1161 & 0.1755 & 0.1609 \\
NPRec\textsubscript{Base} & 0.2146 & 0.2273 & 0.2703 & 0.2429 & 0.2614 & 0.2227 \\
NPRec\textsubscript{Hybrid} & 0.2492 & 0.2473 & 0.3051 & 0.2516 & 0.2873 & 0.2893 \\
\midrule
LLaMA-2-70B & 0.2117 & 0.1999 & 0.1558 & 0.1450 & 0.1816 & 0.1749 \\
NPRec\textsubscript{Hybrid} & 0.2633 & 0.2505 & 0.3146 & 0.2927 & 0.3104 & 0.2921 \\
\bottomrule
\end{tabular}}
  \label{tab:LLM}%
\end{table}

\begin{table}[t] \small
  \centering
    \caption{Case studies of guidelines across different scenarios. Direct LLM denotes guidelines generated by direct prompting. NPRec\textsubscript{w/o debias} refers to guidelines refined by NPRec without the counterfactual debiasing procedure, while NPRec represents the results from the full NPRec framework. The score in brackets indicates the guideline's relative popularity within the global pool, where 100\% corresponds to the most frequent guideline.}
      \vspace{-0.75em}
    \begin{tabular}{l|p{20.835em}}
    \toprule
    \toprule
   \multicolumn{2}{c}{User Profile} \\ 
   \hline
    \multicolumn{2}{c}{Man, around 30 years old, farmer} \\
    \midrule
    \midrule
\multicolumn{2}{c}{User History} \\ 
\hline
    \multicolumn{2}{p{21.835em}}{Home Alone, Philadelphia, Three Musketeers, The, River Wild, The, What's Love Got to Do with It?, Corrina, Corrina, Man Without a Face, The, Wyatt Earp, Free Willy, Sliver} \\
    \midrule
    \midrule
    \multicolumn{1}{c|}{Methods} & \multicolumn{1}{c}{Guidelines} \\ \hline
    Purely LLM & * Films with intense and suspenseful plots.\newline{}* Enjoyment of sports-themed movies. \\
    \midrule
    NPRec\textsubscript{w/o debias} & * Strong Core in 90s Drama. [94\%]\newline{}* Family-Oriented Themes. [79\%]\\
    \midrule
    NPRec & * Emotionally resonant dramas featuring protagonists who must overcome significant social, legal, or personal adversity. [31\%]\newline{}* Family-Oriented Themes. [79\%]\\
    \bottomrule
    \end{tabular}
  \label{tab:guidelinecase}
\end{table}%

\subsection{Guideline Utility and Influence (RQ3)}
To understand the role of guidelines in NPRec, we analyze them from two perspectives: a qualitative case study and a guideline intervention.

\textbf{(1) Case Study.} 
Table \ref{tab:guidelinecase} compares guidelines generated by the base LLM, NPRec without debiasing, and the full NPRec framework.
Purely prompting-based guidelines mainly capture explicit patterns in the user history. They fail to capture latent patterns such as ``Family-Oriented Themes'' that are not immediately evident in the history. In contrast, both NPRec variants identify these latent interests by leveraging semantic correlations across the broader user base. However, without the counterfactual debiasing procedure (NPRec w/o debias), the model tends to retrieve generic, high-frequency guidelines like ``Strong Core in 90s Drama.'' This guideline has a high global frequency (occurring more frequently than 94\% of other guidelines), indicating that the selection is heavily influenced by systemic popularity bias. The full NPRec framework successfully mitigates this bias, retrieving a more granular and personalized guideline which offers a much more precise reflection of the user’s nuanced, intrinsic preferences.

\textbf{(2) Guideline Intervention Analysis.}
A key question for NPRec is whether the retrieved guidelines are functionally involved in recommendation, rather than serving as post-hoc explanations after the ranking has been determined. To examine this, we conduct a controlled guideline intervention analysis that tests the model's semantic responsiveness to the provided guidelines.
For each test case, we keep the user history, candidate item set, and LLM backbone fixed, and only replace the original guideline set with an alternative set of comparable format and length. We then compare the item rankings before and after the intervention. If guidelines are only descriptive, changing them should have little effect on the ranking. In contrast, if they act as part of the reasoning input, modifying them should alter the final recommendation. All results are obtained on the MovieLens dataset using GPT-3.5 as the LLM backbone.

\begin{figure}[t]
  \centering
  \includegraphics[width=0.9\linewidth]{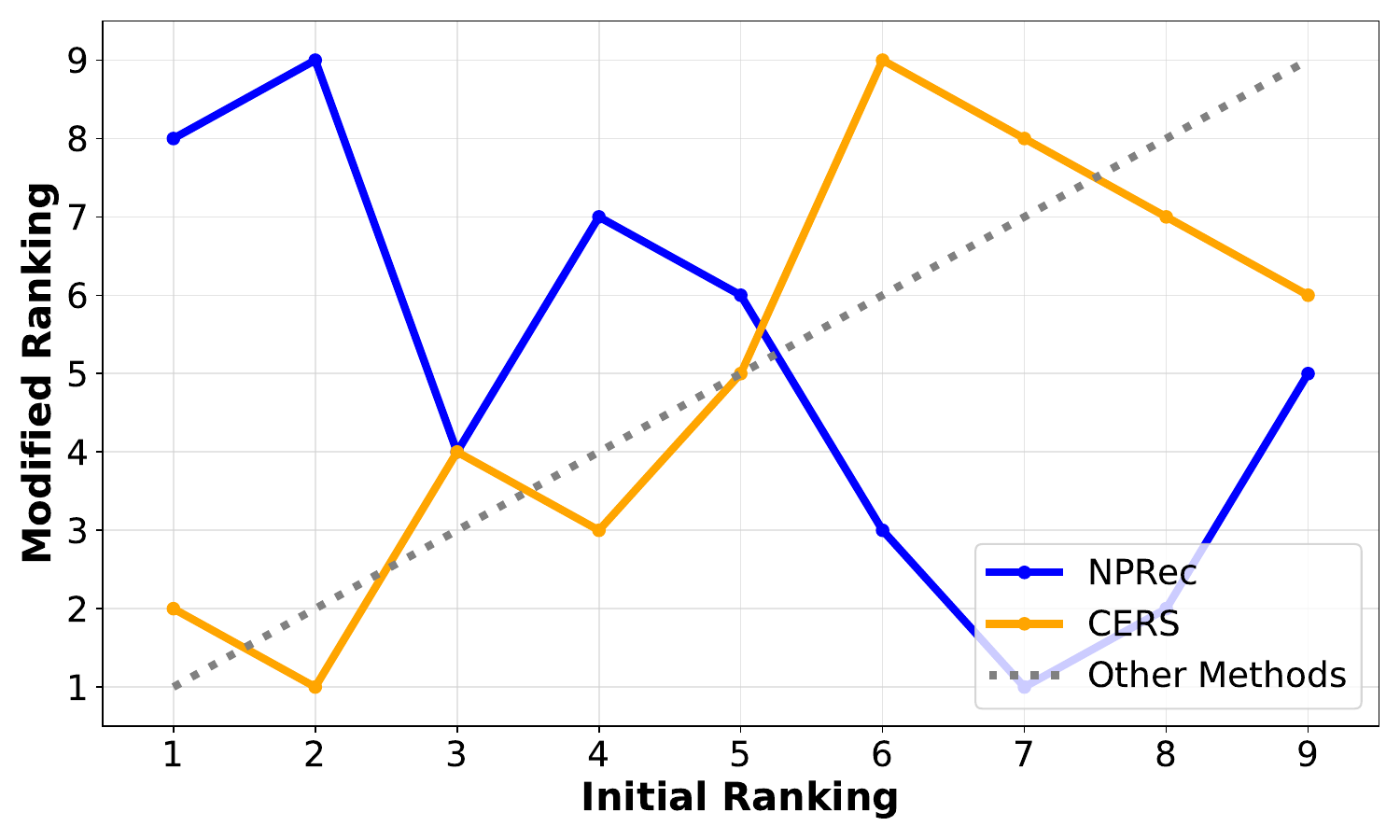}
  \caption{Impact of guideline intervention on item ranking. The axes represent item ranks before (x-axis) and after (y-axis) guideline modification. Deviation from the diagonal ($y=x$) indicates the sensitivity of the final ranking to the provided guidelines.}
          \vspace{-0.75em}
  \label{modi}
\end{figure}

Figure~\ref{modi} shows the rank-transition plot, where the x-axis denotes the original item rank and the y-axis denotes the rank after guideline modification. Points on the diagonal indicate unchanged rankings, while off-diagonal points indicate sensitivity to the modified guidance. As shown in Figure~\ref{modi}, NPRec produces clear off-diagonal transitions, suggesting that its ranking decisions respond to changes in the provided guidelines.
We further quantify this effect using two intervention-based metrics: \textit{Ave. Rank Displacement}, which measures the average absolute change in item rank, and \textit{Top-1 Flip Rate}, which measures the proportion of cases where the top-ranked item changes after intervention. Larger values indicate stronger responsiveness to semantic guidance. The results are reported in Table~\ref{tab:guidelines_invention}. For baselines whose explanations are generated after ranking and are not used as model inputs, we apply the same text-side intervention to the explanation content and re-run the pipeline. Since their recommendation modules do not condition on the explanation text, their rankings remain unchanged, resulting in zero intervention scores. In contrast, NPRec exhibits much larger rank shifts and a higher Top-1 Flip Rate, indicating that its guidelines act as active control signals during recommendation rather than post-hoc attachments.

\begin{table}[t]
  \centering
  \caption{Guideline intervention analysis of different explainable recommendation models. \textit{Ave. Rank Displacement} denotes the average absolute rank change after guideline intervention, and \textit{Top-1 Flip Rate} denotes the proportion of cases in which the top-ranked item changes.}
          \vspace{-0.75em}
  \resizebox{\linewidth}{!}{
    \begin{tabular}{lcccccc}
    \toprule
          & CountER & FGCR  & PEPLER & XRec  & CERS  & NPRec \\
    \midrule
    Ave. Rank Displacement & 0     & 0     & 0     & 0     & 1.95  & 3.67 \\
    Top-1 Flip Rate & 0     & 0     & 0     & 0     & 0.86  & 0.97 \\
    \bottomrule
    \end{tabular}}
  \label{tab:guidelines_invention}
\end{table}

\section{Conclusion and Future Work}
In this work, we investigate and address the critical challenge of popularity bias in LLM-based recommender systems and introduce NPRec, a model-agnostic framework that mitigates this bias not by modifying model parameters, but through external semantic intervention. By externalizing latent bias into tractable textual descriptions and applying counterfactual refinement, NPRec successfully decouples genuine user interests from popularity-driven conformity and provides personalized recommendations with faithful explanations. Our experiments across three real-world datasets demonstrate that NPRec improves recommendation accuracy, provides explanations aligned with the decision process, and mitigates popularity bias. Future work will focus on enhancing guideline precision via external knowledge graphs and multi-agent frameworks, alongside developing human-in-the-loop mechanisms for interactive refinement.

\begin{acks}
This work is supported by the NSFC (No. 62576163, 62376120) and the Fundamental Research Funds for the Central Universities (No. 2026300382).
\end{acks}

\section*{Generative AI Usage Disclosure}
During the preparation of this manuscript, we utilized GPT-5.5 solely for the purpose of language polishing, proofreading, and grammatical refinement to improve the readability of the text. The generative AI tool was not used for any part of the research methodology, experimental design, data collection, or result analysis.

\bibliographystyle{ACM-Reference-Format}
\bibliography{sample-base}

\end{document}